\documentclass[a4paper, 10pt, conference]{ieeeconf}
\usepackage{FG2018}
\usepackage{multirow}
\usepackage{cite}
\usepackage{amsmath,amssymb,amsfonts}
\usepackage{graphicx}
\usepackage{xcolor}
\usepackage{booktabs}
\usepackage{url}
\usepackage{hyperref}
\hypersetup{
  colorlinks=true,
  linkcolor=blue!70!black,
  urlcolor=blue!70!black
}
\FGfinalcopy 

\IEEEoverridecommandlockouts

\def\FGPaperID{****} 

\title{\LARGE \bf
A Multimodal LSTM for Predicting \\ Listener Empathic Responses Over Time
}

\author{\parbox{16cm}{\centering
    {\large Zhi-Xuan Tan$^\dagger$, Arushi Goel$^\dagger$, Thanh-Son Nguyen$^\dagger$, and Desmond C. Ong}\\
    {\normalsize
    A*STAR Artificial Intelligence Initiative, Agency for Science, Technology and Research, Singapore}}
    \thanks{$^\dagger$ Z.X.T, A.G. and T-S.N. contributed equally to this work.}%
    \thanks{This work was supported by the A*STAR Human-Centric Artificial Intelligence Programme (SERC SSF Project No. A1718g0048).}
}

\newcommand{\etal}{\textit{et al}. }

\begin{document}

\ifFGfinal
\thispagestyle{empty}
\pagestyle{empty}
\else
\author{Anonymous FG 2018 submission\\ Paper ID \FGPaperID \\}
\pagestyle{plain}
\fi
\maketitle

\begin{abstract}

People naturally understand the emotions of---and often also \textit{empathize} with---those around them. In this paper, we predict the emotional valence of an empathic listener over time as they listen to a speaker narrating a life story. We use the dataset provided by the OMG-Empathy Prediction Challenge, a workshop held in conjunction with IEEE FG 2019. We present a multimodal LSTM model with feature-level fusion and local attention that predicts empathic responses from audio, text, and visual features. Our best-performing model, which used only the audio and text features, achieved a concordance correlation coefficient (CCC) of .29 and .32 on the Validation set for the Generalized and Personalized track respectively, and achieved a CCC of .14 and .14 on the held-out Test set. We discuss the difficulties faced and the lessons learnt tackling this challenge.

\end{abstract}



\section{Introduction}

    People display emotions all the time in daily life, but more than that, are also sensitive to the emotions of those around them \cite{ong2015affective, ong2018computational}, and often ``feel" what other people are feeling. Watching someone cry with happiness, even through a television or movie screen, would make most of us feel emotional as well \cite{zaki2014empathy, wondra2015appraisal}. Such \emph{empathic} responses are a crucial component of our daily functioning, but have been relatively under-studied in the affective computing literature.
    

Research in building artificial socially-intelligent agents have mainly focused on the person experiencing (and expressing) emotions. Consider Andrew, who is relating an emotional life story to Lisa. Andrew is smiling broadly, tripping over his words, hands gesturing wildly, as he describes his travels in an exotic country. Much of the work in emotion recognition aims to predict Andrew's emotions from one or more modalities: what he is saying, how he is saying it, what his facial expressions convey. 

By contrast, the focus in this paper, based on the OMG-Empathy Prediction Challenge 2019, is not on recognizing emotions in the person expressing them (Andrew). Rather, we aim to recognize the empathic responses of others (Lisa) who are listening to---and \emph{empathizing with}---the emoter. Lisa is engaged with Andrew's story, and can almost feel the same excitement that he has while recounting the story. Here, we present several models that aim to learn Lisa's empathic responses from both Andrew and Lisa's behaviour and the content of Andrew's story, with a challenge dataset of several \textbf{A}ctors narrating stories to several \textbf{L}isteners (see Methods).
    



There has been almost no work in affective computing on predicting empathic responses, so we briefly review work in multimodal emotion recognition, especially using multimodal deep learning \cite{poria2017review,zadeh2016multimodal}. 
%
We drew inspiration from several papers tackling similar challenges. First, W{\"o}llmer \etal \cite{wollmer2013lstm} proposed a system based on Long Short Term Memory (LSTM) long-range temporal context modeling on the Audio/Visual Emotion Challenge (AVEC) dataset. 
Second, a hybrid network with visual attention \cite{guo2018group} was the winning submission for group-level emotion recognition in the Emotion Recognition in the Wild (EmotiW) Challenge 2018. 
%
Third, the OMG-Emotion Behaviour Challenge 2018 \cite{Barros2018OMG} produced several models, such as a bidirectional LSTM \cite{peng2018omg} and utterance-level sentiment regression \cite{deng2018multimodal}, that predict a speaker's valence from short one-minute videos. 



In this paper, we propose a multimodal LSTM model with a local attention layer for predicting the valence ratings of the listener in the OMG-Empathy Dataset.
Our models are available as a public GitHub repository\footnote{\href{https://github.com/desmond-ong/cheem-omg-empathy}{\small {https://github.com/desmond-ong/cheem-omg-empathy}}}.

\section{Methods}

\subsection{Dataset}

We use the OMG-Empathy Dataset released with this challenge, 
which consists of semi-scripted interactions between pairs of individuals, one of whom is an actor, and the other, a na\"ive participant (or ``listener''). The actor recounts a fictional autobiographical story to the listener. In total, there were 4 actors, each of whom told 2 stories, such as ``talking about a childhood friend", or ``I had a bad flight experience". Ten participants were recruited as listeners, and each listener listened to every story. This gives a total of 80 videos, each lasting on average 5 minutes 12 seconds.
The dataset was split into Training (4 stories, 40 videos), Validation (1 story, 10 videos), and Test (3 stories, 30 videos) sets.

After each session, listeners were shown a video of their interactions, and were asked to rate using a joystick how they themselves felt as the actor was telling the story (i.e., their empathic responses). These annotations were continuous in time, and ranged from -1 (negative) to 1 (positive).

\subsection{Data Augmentation}

Given the limited amount of training data, we opted to split each $5$-min training video into five $1$-min segments, augmenting the number of training sequences from $40$ to $200$, and allowing us to better train our models using batched gradient descent. While this breaks the standard assumption that training examples are conditionally independent given the generating process, we found empirically that this was made up for by the increase in training data. The effectiveness of this approach suggests that temporal dependencies over the $1$-min time-scale are not crucial for predicting empathy.

\subsection{Feature Extraction}

\subsubsection{Audio Features} We used openSMILE v2.3.0 \cite{eyben2013opensmile} with the accompaning \texttt{emobase} configuration file to extract 990 low-level acoustic features for every 1-second window.

\subsubsection{Text Features} We used YouTube's automatic subtitling to extract transcripts along with start and end timings of individual ``utterances''\footnote{An ``utterance'' is not necessarily a complete sentence.}. We used 300-dimensional GloVe word embeddings \cite{pennington2014glove} to obtain features per utterance by averaging across the embeddings of its words. For synchronization with other modalities, the utterance feature was repeated for each 1-second window within the timespan of the utterance, with the most recent utterance chosen if a window overlapped with multiple utterances. 

\subsubsection{Visual Features} We extracted the fully-connected 4096-dimensional feature embeddings from the pre-trained VGG Face CNN descriptor models \cite{Parkhi15}. The features are extracted for every frame from the listener's face and then averaged over a period of 1-second (25 frames). We also extracted the actor's facial features, but found that adding actor features to our models usually hurt their performance.




\subsection{Long Short Term Memory (LSTM) Models}

\begin{figure}[ht]
  \centering
  \includegraphics[width=.90\columnwidth]{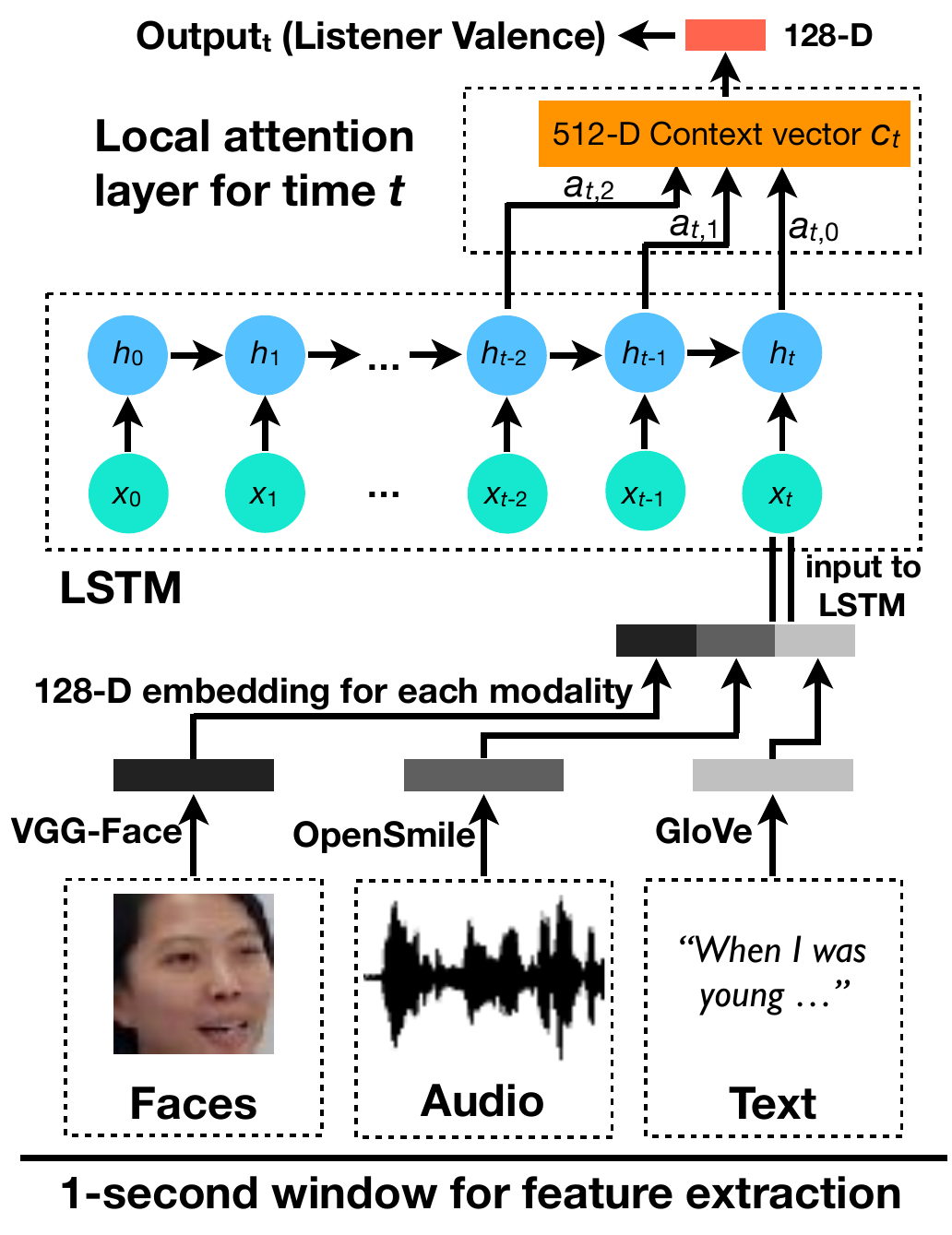}
  \caption{Model architecture. We extract multimodal features $x_t$ for every 1-second window, which is input into an LSTM. (We omit the inner LSTM operations for brevity). We introduce a local attention layer that weights contributions from $h_t$ up to $h_{t-2}$ using attention weights $a_{t,0},\ldots,a_{t,2}$. The final output layer is a regression to predict the listener's valence rating.}
  \label{fig:model}
\end{figure}

Our LSTM model architecture is shown in Fig. \ref{fig:model}. We first extract multimodal features for each 1-second window, compress them to a 128-dimensional embedding for each modality and concatenate the feature vectors (feature-level fusion). These feature vectors are then fed into a LSTM recurrent neural network \cite{hochreiter1997long}, allowing information from previous time points to influence the hidden layer representation at time $t$. During training, we ran the LSTM over the 1-min video segments described in the Data Augmentation section, while at validation and test time, we used the entire video.

Next, we introduced a local attention layer \cite{luong2015effective} which weights contributions from a small local window $[t-(W-1),t]$ around the current time-point $t$. The attention layer outputs a context vector $c_t$ by computing a weighted sum over the hidden states of the LSTM within the window:
\begin{equation}
    c_t = \sum_{i=0}^{W-1} a_{t,i} \cdot h_{t-i}
\end{equation}
Here, $a_{t,\cdot}$ is a vector of $W$ attention weights, $h_t$ is the hidden state of the LSTM, and $W=3$ seconds is the window size, which was selected through cross-validation. The attention weights are computed as a function $a_{t,\cdot} = f_\phi(x_t)$ of the multimodal inputs $x_t$, where $f_\phi(\cdot)$ is parameterized by the weights $\phi$ of a multi-layer perceptron with a 128-dimensional hidden layer. The model is thus able to learn which hidden states over the recent past matter most when predicting the valence. Finally, the 512-dimensional context vector $c_t$ is fed into a 128-dimensional layer, then into a linear regression that predicts the listener's valence rating.

We experimented with multiple models using the same basic architecture, varying the input modalities and other design choices. For our text-only model (T), we opted to run our LSTM over sequences of utterances instead of 1-second windows, and found that predicting the change in the valence, rather than the valence itself, led to less over-fitting and better performance. For the text-only (T) and vision-only (V) models, we also omitted the attention layer, finding that it did not improve performance. In total, we evaluated 7 different LSTM models: unimodal (A: audio, T: text, V: visual), bimodal (AT, AV, TV), and trimodal (ATV).



\section{Results}

To evaluate our results, we use the Concordance Correlation Coefficient (CCC) between the model's predictions and participants' self-reported empathic responses. The CCC is similar to the Pearson correlation coefficient $\rho$, in that it measures the agreement between two vectors, where $+1$ corresponds to perfect agreement. Unlike $\rho$, the CCC also takes into account the difference between the means of both vectors, penalizing bias in the model's predictions.

\subsection{Cross-Validation Results}

In our initial experiments, we realized that the stimuli were very heterogeneous---the same model can perform very well on some stories, and very poorly on other stories. Thus, we also prioritized finding models (modality-combinations and hyperparameters) that generalize well across different stories.

To do this, we performed leave-one-out cross-validation across the provided stories, training on four of the stories and computing the CCC for the held-out story. We then computed the mean and standard deviation of the CCC across stories, reported in Table \ref{tab:cvCCC}, to evaluate the models' robustness to variations in the training data. The first three models listed are unimodal LSTMs, the next four are multimodal LSTMs that employ early feature fusion (as in Fig. \ref{fig:model}). 


\begin{table}[h]
\caption{CCC values cross-validated across stories}
\label{tab:cvCCC}
\centering
\begin{tabular}{@{}crrrrrl@{}}
\toprule
\multirow{3}{*}{\textbf{Model}} & \multicolumn{6}{c}{\textbf{CCC}} \\ \cmidrule(l){2-7} 
 & \multicolumn{5}{c}{\textit{Validation Story}} & \multicolumn{1}{l}{\multirow{2}{*}{\textit{Mean} $\pm$ \textit{SD}}} \\
 & \multicolumn{1}{l}{\textit{1}} & \multicolumn{1}{l}{\textit{2}} & \multicolumn{1}{l}{\textit{4}} & \multicolumn{1}{l}{\textit{5}} & \multicolumn{1}{l}{\textit{8}} & \multicolumn{1}{l}{} \\ \midrule
A & .22 & -.10 & .14 & .15 & .00 & .08 $\pm$ .12 \\
T & .19 & .00 & \textbf{.27} & .13 & .12 & \textbf{.14} $\pm$ .10 \\
V & .04 & \textbf{.08} & -.14 & .10 & \textbf{.19} & .05 $\pm$ .12 \\ \midrule
AT & \textbf{.26} & -.04 & .11 & .12 & .05 & .10 $\pm$ .10 \\
AV & .18 & -.22 & .16 & .15 & .01 & .05 $\pm$ .15 \\
TV & .02 & -.04 & .01 & .07 & .16 & .04 $\pm$ .07 \\
ATV & .19 & -.08 & .15 & \textbf{.18} & .10 & .11 $\pm$ .10 \\ 
\bottomrule
\end{tabular}
\end{table}

Interestingly, most of our models performed best when validated on the original Validation set (Story 1), but very poorly when validated on Story 2, which often resulted in negative CCC values. As a result, the mean cross-validated CCC was substantially lower than the CCC for the best story, and the standard deviation tends to be almost as large as, if not larger than, the mean. While disappointing, given the consistency of this trend across models, one hypothesis is that Story 2 is an outlier in comparison to the remaining four stories. The only model that performed reasonably on Story 2 was the visual-only model (V), but unfortunately it performed poorly on Stories 1 and 4 instead.

Another notable trend was that increasing the number of modalities did not reliably improve performance across stories. The audio-text model (AT) performed best on the original Validation set, but performed worse on cross-validation than the text-only model (T). The audio-text-visual model (ATV) performed better on cross-validation than AT, but worse on the original Validation set. 

\subsection{Generalized Track}

Informed by our cross-validation results, our three submissions for the \textbf{G}eneralized Empathy track are as follows:

\vspace{0.25em}
\noindent \begin{tabular}{lrl}
    {[}\textbf{G1}{]}: & AT   & Best model on Validation Set (Story 1) \\
    {[}\textbf{G2}{]}: & T    & Best model on cross-validation \\
    {[}\textbf{G3}{]}: & ATV  & Best multimodal model on cross-validation
\end{tabular}
\vspace{0.05em}

We selected models using these criteria based on different assumptions about the representativeness of the training set. 
For [\textbf{G1}], we chose the best-performing model (AT) on the original Validation set (Story 1). 
For [\textbf{G2}], we chose the model with the best cross-validation performance; this happened to be the text-only model (T). Finally, for [\textbf{G3}], we chose the best-performing multimodal model (ATV) to see if multimodality would improve generalization.
We note that even though the modality combinations in [\textbf{G2}] and [\textbf{G3}] were selected via leave-one-out cross-validation, the submission models are still only trained on the Training set.

The validation CCC of our submissions can be compared against the challenge baseline in the second column of Table \ref{tab:submitCCC}. We also report the standard deviation across videos as an error estimate. For submission purposes, we used early-stopping to determine the best epoch for each model on the original validation set (Story 1). This explains the difference from the Story 1 column in Table \ref{tab:cvCCC}, which shows the results for the epoch that gives the best cross-validated CCC. It can be seen that our models substantially outperform the baseline provided by the organizers, with the audio-text model [\textbf{G1}] achieving a CCC of $.29$ with a standard deviation of .12 (calculated across the videos in the Validation set).


\subsection{Personalized Track}

To address the \textbf{P}ersonalized Track of the challenge, which requires the model to predict the empathic responses of specific listeners, we took our Generalized models [\textbf{G1}--\textbf{G3}] as pre-trained seed models, and fine-tuned them for each listener to produce models [\textbf{P1}--\textbf{P3}]. Fine-tuning meant training only on videos from one specific listener for an additional 250 epochs, with early-stopping to select the best epoch. This fine-tuning yielded ten personalized models, one for each of ten listeners. The CCC values on the Validation set (averaged across listeners) are displayed in the fourth column of Table \ref{tab:submitCCC}. As expected, fine-tuning to each listener improved upon the generalized models, with the ATV model improving from a CCC of $.23$ to $.28$. Similar to the Generalized track, our Personalized models also substantially outperformed the provided baseline on the Validation set.


\begin{table}[h]
\caption{CCC values for Submitted Models (Mean $\pm$ Std. Dev.)}
\label{tab:submitCCC}
\centering
\begin{tabular}{@{}lllll@{}}
\toprule
\multirow{3}{*}{\textbf{Model}} & \multicolumn{4}{c}{\textbf{CCC}} \\ \cmidrule(l){2-5} 
 & \multicolumn{2}{c}{\textit{Generalized}} & \multicolumn{2}{c}{\textit{Personalized}} \\
 & \multicolumn{1}{c}{Valid.} & \multicolumn{1}{c}{Test} & \multicolumn{1}{c}{Valid.} & \multicolumn{1}{c}{Test} \\ \midrule
Baseline & .11 & -- & .09 & -- \\
{[}\textbf{G1/P1}{]} AT & \textbf{.29} $\pm$ .12 & \textbf{.14} $\pm$ .19 & \textbf{.32} $\pm$ .11 & \textbf{.14} $\pm$ .20 \\
{[}\textbf{G2/P2}{]} T & .19 $\pm$ .11 & .11 $\pm$ .21 & .20 $\pm$ .12 & .07 $\pm$ .20 \\
{[}\textbf{G3/P3}{]} ATV & .23 $\pm$ .12 & .07 $\pm$ .15 & .28 $\pm$ .13 & .07 $\pm$ .19 \\ \bottomrule
\end{tabular}
\end{table}

   \begin{figure}[tb]
      \centering
      \includegraphics[width=.7\columnwidth]{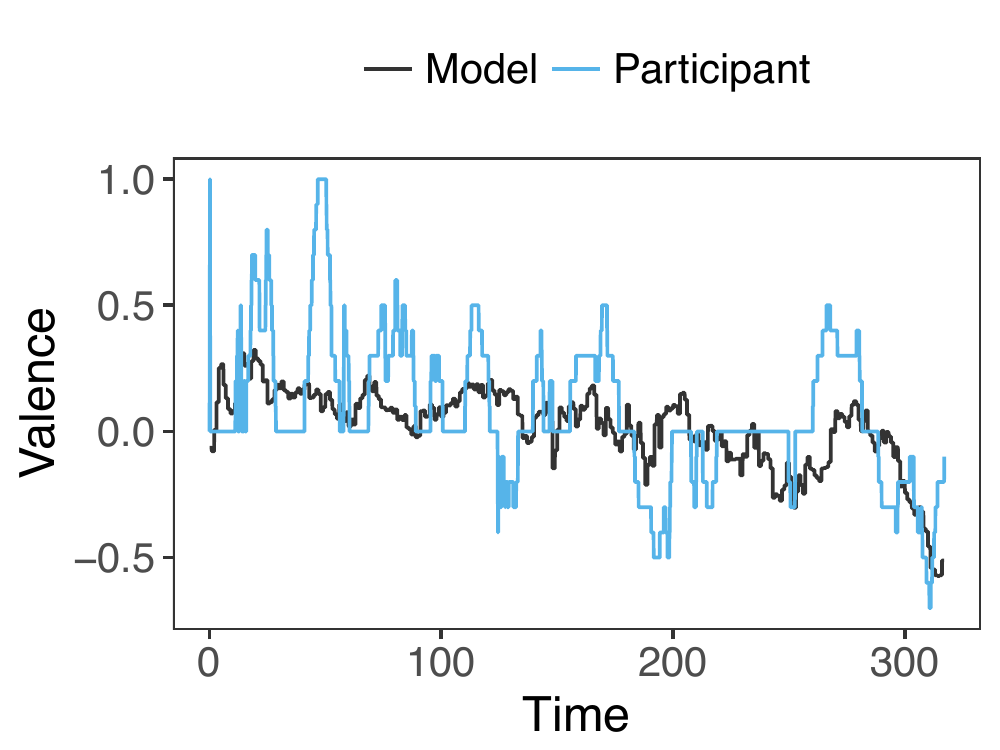}
      \caption{Output of {[}\textbf{G3}{]} on Subject 2's ratings on the Validation Story. The model tracks the ratings well during the latter half of the video.}
      \label{fig:exampleOutput}
   \end{figure}

\subsection{Test Performance}

Unfortunately, our models did not perform as well on the Test set, as shown in the `Test columns of Table \ref{tab:submitCCC}. Our best Generalized and Personalized models ([\textbf{G1}] and [\textbf{P1}]) both achieved a CCC of $.14$. Nonetheless, this still outperformed the validation baselines, and fell within our estimated error bounds in the last column of Table \ref{tab:cvCCC}. Much of this decrease in performance can be traced to poor performance on Story 7 in the test set, which had CCC values of $.02$, -$.07$ and -$.05$ for models [\textbf{G1}] to [\textbf{G3}] respectively. As was indicated by our poor cross-validation performance on Story 2, inter-story variation can make robust empathy prediction a difficult task. This also explains the increase in standard deviation over videos in the test set---our models were not robust enough for the videos in Story 7. The personalized models did not outperform the generalized models despite fine-tuning, suggesting that fine-tuning led to over-fitting rather than personalization to each listener.

\section{Discussion}

When empathizing with someone as they tell a story, perhaps the most important information that influences one's responses is the \emph{content of their story}. This intuition was borne out in our experiments, which showed that models that include the text modality do better than models without. Indeed, we only scratched the surface of the textual information available: We used word-level features, ignoring word order in each utterance. These design choices were compromises given the time and data constraints of the Challenge---transcripts were not provided, so we used YouTube-generated subtitles that were not segmented by speaker or sentence---which in turn may have hurt the performance of our models. Given more time, a more principled approach would account for the syntactic structure of each utterance, as well as model the fact that utterances in this dataset occur as a dialogue. Segmenting the text and audio modalities by speaker would also have allowed separate affect models to be built for actor and listener, which might have better captured the influence of the actor upon the listener's empathic responses.

Another challenge we encountered was the heterogeneity of the stories. First, the stories all express very different emotions, and there may not have been enough examples for our models to learn generalizable features. Second, as noted in the Challenge description, each story was told by a particular actor intentionally expressing a particular personality trait. This resulted in considerable variance in model performance across stories. For example, on the Test set, our models did extremely well on Story 6 (e.g., [\textbf{G2}]: .26) but poorly on Story 7 ([\textbf{G2}]: -.07).



One final challenge of siginficance was the nature of the ratings and the scale used. See Fig. \ref{fig:exampleOutput} as an example: Participants' ratings were highly discontinuous, with large, sharp changes. Our models tended to make smaller changes more frequently, generally capturing the trend but not the larger discontinuities and extremities. Future work could adapt the model architecture to deal with sharper changes, perhaps by predicting both the rating and the rate of change in the rating (i.e. the kinematics of valence), or by modelling valence changes as event-driven \cite{qin2017event}.

In conclusion, much work remains to be done in building models that can reason about an individual's emotions, but also the emotions of others interacting with that individual. There are many challenges to be solved, from the complexity of rich narrative stimuli to dealing with time-series data. We hope that this challenge will inspire future research into modelling empathic responses over time, and will pave the way for building applications like empathic robots and digital personal assistants.



\bibliographystyle{ieee}
\bibliography{omgEmpathy}

\begin{thebibliography}{10}\itemsep=-1pt

\bibitem{Barros2018OMG}
P.~Barros, N.~Churamani, E.~Lakomkin, H.~Sequeira, A.~Sutherland, and
  S.~Wermter.
\newblock {The OMG-Emotion Behavior Dataset}.
\newblock In {\em 2018 International Joint Conference on Neural Networks
  (IJCNN)}, pages 1408--1414. IEEE, 2018.

\bibitem{deng2018multimodal}
D.~Deng, Y.~Zhou, J.~Pi, and B.~E. Shi.
\newblock Multimodal utterance-level affect analysis using visual, audio and
  text features.
\newblock {\em arXiv preprint arXiv:1805.00625}, 2018.

\bibitem{eyben2013opensmile}
F.~Eyben, F.~Weninger, F.~Gross, and B.~Schuller.
\newblock Recent developments in opensmile, the munich open-source multimedia
  feature extractor.
\newblock In {\em Proceedings of the 21st ACM International Conference on
  Multimedia}, pages 835--838. ACM, 2013.

\bibitem{guo2018group}
X.~Guo, B.~Zhu, L.~F. Polan{\'\i}a, C.~Boncelet, and K.~E. Barner.
\newblock Group-level emotion recognition using hybrid deep models based on
  faces, scenes, skeletons and visual attentions.
\newblock In {\em Proceedings of the 2018 on International Conference on
  Multimodal Interaction}, pages 635--639. ACM, 2018.

\bibitem{hochreiter1997long}
S.~Hochreiter and J.~Schmidhuber.
\newblock Long short-term memory.
\newblock {\em Neural Computation}, 9(8):1735--1780, 1997.

\bibitem{luong2015effective}
T.~Luong, H.~Pham, and C.~D. Manning.
\newblock Effective approaches to attention-based neural machine translation.
\newblock In {\em Proceedings of the 2015 Conference on Empirical Methods in
  Natural Language Processing}, pages 1412--1421, 2015.

\bibitem{ong2015affective}
D.~C. Ong, J.~Zaki, and N.~D. Goodman.
\newblock Affective cognition: Exploring lay theories of emotion.
\newblock {\em Cognition}, 143:141--162, 2015.

\bibitem{ong2018computational}
D.~C. Ong, J.~Zaki, and N.~D. Goodman.
\newblock Computational models of emotion inference in theory of mind: A review
  and roadmap.
\newblock {\em Topics in Cognitive Science}, 2018.

\bibitem{Parkhi15}
O.~M. Parkhi, A.~Vedaldi, and A.~Zisserman.
\newblock Deep face recognition.
\newblock In {\em British Machine Vision Conference}, 2015.

\bibitem{peng2018omg}
S.~Peng, L.~Zhang, Y.~Ban, M.~Fang, and S.~Winkler.
\newblock {A Deep Network for Arousal-Valence Emotion Prediction with
  Acoustic-Visual Cues}.
\newblock In {\em arxiv}, 2018.

\bibitem{pennington2014glove}
J.~Pennington, R.~Socher, and C.~D. Manning.
\newblock Glove: Global vectors for word representation.
\newblock In {\em Empirical Methods in Natural Language Processing (EMNLP)},
  pages 1532--1543, 2014.

\bibitem{poria2017review}
S.~Poria, E.~Cambria, R.~Bajpai, and A.~Hussain.
\newblock A review of affective computing: From unimodal analysis to multimodal
  fusion.
\newblock {\em Information Fusion}, 37:98--125, 2017.

\bibitem{qin2017event}
Z.~Qin and C.~R. Shelton.
\newblock Event detection in continuous video: An inference in point process
  approach.
\newblock {\em IEEE Transactions on Image Processing}, 26(12):5680--5691, 2017.

\bibitem{wollmer2013lstm}
M.~W{\"o}llmer, M.~Kaiser, F.~Eyben, B.~Schuller, and G.~Rigoll.
\newblock Lstm-modeling of continuous emotions in an audiovisual affect
  recognition framework.
\newblock {\em Image and Vision Computing}, 31(2):153--163, 2013.

\bibitem{wondra2015appraisal}
J.~D. Wondra and P.~C. Ellsworth.
\newblock An appraisal theory of empathy and other vicarious emotional
  experiences.
\newblock {\em Psychological Review}, 122(3):411, 2015.

\bibitem{zadeh2016multimodal}
A.~Zadeh, R.~Zellers, E.~Pincus, and L.-P. Morency.
\newblock Multimodal sentiment intensity analysis in videos: Facial gestures
  and verbal messages.
\newblock {\em IEEE Intelligent Systems}, 31(6):82--88, 2016.

\bibitem{zaki2014empathy}
J.~Zaki.
\newblock Empathy: A motivated account.
\newblock {\em Psychological Bulletin}, 140(6):1608, 2014.

\end{thebibliography}

\end{document}